\documentstyle[11pt,fullpage]{article}

\title{From compositional to systematic semantics}
\author{ Wlodek Zadrozny \\
IBM Research, T. J. Watson Research Center \\
Yorktown Heights, NY  10598  \\
{\sc WLODZ @ WATSON.IBM.COM}
\thanks{Linguistics and Philosophy(17):329-342, 1994}
}
\date{}

\begin{document}
\bibliographystyle{named}
\maketitle

\begin{abstract}

We prove a theorem stating that any semantics can be encoded as a
compositional semantics, which means that, essentially, the standard
definition of compositionality is formally vacuous. We then show that when
compositional semantics is required to be "systematic"
(that is, the meaning
function cannot be arbitrary, but must belong to some class),
it is possible to
distinguish between compositional and non-compositional semantics.
As a result, we believe that the paper clarifies the
concept of compositionality and opens a possibility of making systematic
formal comparisons of different systems of grammars.
\end{abstract}

\section{Introduction}

Compositionality is defined as the property that the
meaning of a whole is a function of the meaning of its parts (cf. e.g.
\cite{KeenanandFaltz85}, pp.24-25). (A slightly less general definition,
e.g. \cite{Partee90}, postulates the existence of a
homomorphism from syntax to semantics). However,
we can prove a theorem stating that any semantics can be encoded as a
compositional semantics, which means that, essentially, the standard
definition of compositionality is formally vacuous.
Thus, although intuitively clear, the definition is not restrictive
enough. We illustrate the power of the theorem by showing how
to assign compositional semantics to idioms and to a very
counterintuitive semantics of coordination (Section 4).

Given a class of functions $F$, we say that the
compositional semantics is {\em systematic}  if the
meaning function belongs to the class $F$. We show that when
compositional semantics is required to be {\em systematic}
we can distinguish between grammars with
compositional and non-compositional semantics;
we present an example of a simple grammar for which there is no
"systematic" compositional semantics (Section 3).

As a result, we believe that the paper clarifies the
concept of compositionality and opens a possibility of making systematic
comparisons of different systems of grammars and
natural language understanding programs.
Furthermore, the concept of systematicity that we introduce in
the paper might be useful in extracting the formal meaning behind
various versions of compositionality as a philosophical principle,
cf.\cite{Partee82}.
But in this paper we restrict ourselves to the mathematics of
compositionality.

Compositional semantics is usually defined  as  a functional
dependence of the meaning of an expression on the meanings of its  parts.
One   of  the  first  natural  questions  we  might want to ask is whether a
set of natural language expressions, i.e. a language,
can have some compositional semantics. This question has
been answered positively by
\cite{vanBenthem82}. However his result says nothing
about what kinds of things should be assigned e.g. to nouns, where,
obviously,
we would like nouns to be mapped into sets of entities, or something like
that. That is, we want semantics to encode some basic
intuitions, e.g. that
nouns denote sets of entities, and verbs denote relations between entities,
and so on; in other words, we would like to have
a compositional semantics that agrees with intuitions.
More formally, the questions is whether after deciding what
sentences and
their parts mean, we can find a function that would compose the meaning  of a
whole from the meanings of its parts.

The answer to this question is somewhat disturbing. It turns out  that
whatever we decide that some language expressions should mean, it is always
possible to produce a function that would give compositional semantics
to it (see below for a more
precise formulation of this fact). The upshot is that compositionality, as
commonly defined, is not a strong constraint on a semantic theory.

\section{Proving the existence of compositional semantics}

Let $S$  be  any collection of expressions (intuitively, sentences and their
parts). Assume that elements of $S$ (e.g. $s.t$)
are composed from other elements of
$S$ (that is, $s$ and $t$)  by concatenation (".").
We do not assume that concatenation is associative, that is
$ (a.(b.c)) = ((a.b).c) $. Intuitively, this means that we assign
semantics to parse trees,  and not to strings of words.

Let $M$ be a set  of meanings, and let for any $s  \in S$,  $m(s) \in
M$ be the meaning of $s$. We want to show that there
is  a compositional semantics for $S$ which agrees with the function $m$
associating $s$ with $m(s)$.

Since elements of $ M$
can be of any type, we do not automatically have
$ m(s.t)  =  m(s)\sharp m(t) $, where $ \sharp $  is some
operation on the meanings. To
get that kind of homomorphism we have to perform a type raising operation
that would map elements of $S $
into functions  and  then the functions into the
required meanings.
Note that such a type raising operation is quite common
both in mathematics  (e.g. $1$ being a function equal to $1$ for all
values) and in mathematical linguistics.
The  meaning function $\mu$ that we want to define will provide
compositional semantics for $S$ by mapping it into
a set of functions in such a way that $\mu(s.t) = \mu(s) ( \mu(t) ) $,
for all elements $s.t$ of $S$.

Secondly, we want that the original semantics be
easily decoded from  $ \mu(s)$. There is more than one way of
doing this.
One can trivially extend the language $S$ by adding
to   it an "end of expression" character $\$$,
which may appear only as the
last element of any expression. The purpose of it is to encode the function
$m(x)$ in the following way:
$ \mu(s.\$) =  m(s) $, for all $s$ in $S$. Intuitively, the
character $\$$ is like the period at the end of a sentence, or the pause
marking the end of an utterance. In effect, we will be treating all
sentences as idioms, or garden path sentences, where the meanings are
clear only once the sentence is completed (Theorem 2).
But, as we are going to show now, the original
semantics can be encoded in a different way, without extending the original
language, e.g. by assuring $ \mu(s)(s) =  m(s) $,
for all $ s $ in $ S $  (Theorem 1).

To make the notation simple, we have assumed that there is only one
way of composing elements  of S, by concatenation,
but  all our arguments work for languages with many operators as well.
We show an example of how such operators can be handled in Section 4.\\

{\sc {\bf THEOREM 1.}}
Let $M$ be an arbitrary set.
Let $A$ be an arbitrary alphabet. Let "." be a binary operation,
and let $S$ be the set closure of $A$ under ".". Let $m : S
\rightarrow M$  be an arbitrary function.

Then there is a set of functions $M^*$ and a unique
map $\mu : S \rightarrow M^*$ such that for all $s$, $t\in S$
$$\mu(s.t) = \mu(s) ( \mu(t) ),  \ \mbox{ and } \
\mu(s)(s) = m(s) . $$ \\

{\sc {\bf Corollary 2.}}  Theorem 1 is also true when
the binary operation "." is partial.\\
\vspace*{.1in}\\
{\sc {\bf Preliminaries to the proof: The solution lemma}} \\
Our results will be proved in set theory with the
anti-foundation axiom.
This set theory, ZFA,  is equiconsistent with
the standard system of ZFC, thus the theorem does not assume
anything more than what is needed for "standard mathematical
practice". Furthermore, ZFA is better suited as foundations for
semantics of natural language  than ZFC
(\cite{BarwiseEtch87}).

We need only one (but fundamental) theorem of ZFA:
the solution lemma (\cite{Aczel87} and
\cite{BarwiseEtch87}), which says any (well-formed)
collection of equations that define sets
has a unique solution.
For the reader who is not familiar with set theory, the meaning of the
solution lemma can be explained as follows:
We have a universe of sets $V$, and
a set of variables $X= \{  x_1 , x_2 , ... \} $,
which may be infinite (countable or uncountable). We can form equations
of the form
$$   x_i = a\_set(X, V) $$
where $a\_set(X,V)$ is a set expression involving the variables and
elements of $V$, for instance, if $a \in V$ and
$b \in V$, we can write the following equations:
\begin{eqnarray*}
x_7 & = & \{ \{ a \} , x_7 , x_9 \} \\
x_8 & = &
\{ b , \{ a , x_8 \} , \{ \{ \{ x_{79} \} \} \}  \} \\
x_9 & = & \{ b \} \\
x_{79} & = & \{ x_{79} \}
\end{eqnarray*} \\
We say that such a set is well-formed if each variable appear only
once on the left, and each left hand side is a variable.
The solution lemma says that any set of such equations (finite or
infinite) has a unique
solution. That is, there is a unique
collection of sets that satisfy them.
\mbox{ } \hfill $\Box$
\vspace*{.1in} \\
{\sc Proof of Theorem 1 and Corollary 2} \\

{\bf Proof of Theorem 1.}
It is enough to ensure that for all $s \in S$
$$ \mu(s) =  \{<s,m(s)>\}
\ \ \cup \ \ \{ <\mu(t) ,\mu(s.t)>\ :\ t\in S\}\ $$
Clearly,
$\mu$ is a function, because it is a collection of pairs.
The proof is complete once we check that
for  $s$'s and   $ t  $'s in $S$ we have
(i)  $\mu(s . t ) = \mu(s ) ( \mu(t )) $ , and
(ii) $\mu(s)(s) = m(s ) $ .
Using the above equality we check (i): If $f=\mu(s )$, then
$f(\mu(t )) = \mu(s . t ) $. Similarly for (ii).

It remains to show that using the solution lemma we can make the
above equation true for all $s \in S$. We begin by
introducing a set variable
$X_s$ for every  $s \in S$, and observing that
$$ X_s =  \{<s,m(s)>\}
\ \ \cup \ \ \{ <X_t ,X_{s.t}>\ :\ t\in S\}\ $$
is a well-formed set equation for any $s \in S$.
(The pair $< a , b >$ is set theoretically defined as
$ \{ \{ a \} , \{ a , b \} \})$.
Hence the solution
lemma applies, and there are unique sets $\mu(s)$ that satisfy
each equation. But each such $\mu(s)$ is a collection of pairs, i.e.
a function. Furthermore, since each $\mu(s)$ is unique, and $S$ is
a set, the mapping $\mu$ associating $\mu(s)$ with each $s \in S$ is
a function. This completes the proof of Theorem 1. \\
\mbox{ } \hfill $\Box$\\

{\bf Proof of Corollary 2.}
It is enough to observe that we can add an extra condition in the
main equation of Theorem 1, and the proof still works:
$$ \mu(s) =  \{<s,m(s)>\}
\ \ \cup \ \
\{ <\mu(t) ,\mu(s.t)>\ :\ t\in S \mbox{ and }  \ \ s.t \in S
\}\ $$
\mbox{ } \hfill $\Box$\\

{\sc note}. Notice that we can view using the solution lemma in the
above proofs as
an extreme example of defining a function by cases. To see it more
clearly, one can make the main equation of the proof of Theorem 1
explicit.
Let  $ t(0) , t(1) ,   \ldots  , t(\alpha) $
enumerate  $ S $.
We can
create a big table specifying meaning values for all strings and  their
combinations.  Then the  conditions on the meaning functions $\mu(s)$
can be written as the set of equations below
\begin{eqnarray*}
    \mu(t(0)) &  = &  \{   <  t(0) ,  m(t(0) )  >  ,
                    <  \mu(t(0)), \mu(t(0) . t(0) )  > ,
                      \ldots  , \\
          & &       <  \mu(t(\alpha)),  \mu(t(0).t(\alpha))  >  ,  \ldots
                  \}  \\
    \mu(t(1)) & =  & \{   <  t(1) ,  m(t(1) )  >  ,
                    <  \mu(t(0)), \mu(t(1) . t(0) )  > ,
                      \ldots  , \\
       &  &         <  \mu(t(\alpha)),  \mu(t(1).t(\alpha))  >  ,  \ldots
                  \}  \\
     \ldots  \\
      \mu(t(\alpha)) & =  & \{   <  t(\alpha) ,  m(t(\alpha) )  >  ,
                    <  \mu(t(0)), \mu(t(\alpha).t(0) )  > ,
                      \ldots  , \\
       & &          <  \mu(t(\alpha)),  \mu(t(\alpha).t(\alpha))  >  , \ldots
                  \}  \\
   \ldots  \\
\end{eqnarray*}
In ordinary mathematics, this would correspond to saying that if
$x$ is $1$ then $f(x)=32$,
if $x$ is $2$ then $f(x)=14732$,
if $x$ is $3$ then $f(x)=1$, and so on. Clearly, such a process
defines the
function $f$, but, intuitively, it is not a definition we would
care much for. Before showing that requiring a better description
of an $f$ than as a set of pairs makes sense, we want to observe that
the encoding of the original meaning function
can be uniform in the following sense: \\

{\sc {\bf PROPOSITION 3.}}
In addition to the assumptions of Theorem 1, let $\$ \notin  A$, and
let $S\$$ be the language obtained by the mapping
$ s \rightarrow s.\$$, for all $s \in S$.
Then there is a set of functions $M^*$ and a unique
map $\mu : S\$ \rightarrow M^*$ such that for all $s$, $t\in S$
$$ \mu(s.t) = \mu(s) ( \mu(t) ) , \mbox{ and }
\mu(s.\$) = m(s) . $$ \\

Proof. As in the proof of
Corollary 2, we can change the set of equations to
$$ \mu(s) =  \{<\$,m(s)>\}
\ \ \cup \ \ \{ <\mu(t) ,\mu(s.t)>\ :\ t\in S\}\ $$
To finish the construction of $\mu$,
we make sure that the equation
$  \mu(\$) = \$  $
holds. Formally, this requires adding the pair $ < \$ , \$ > $ into
the graph of $\mu$ that was obtained from the solution lemma.
Also, we have to extend the domain of function $\mu$ to include
$S\$$. This is easily done by adding to the already constructed
part of $\mu$ the set of pairs
$ \{ < s.\$ , m(s) > : s \in S \} $.
The proof is complete once we check that for  $s$'s  in $S$ we have
$\mu(s . \$ ) = m(s ) $, and that
$\mu(s . \$ ) = \mu( s ) ( \mu(\$) ) $
(because $ \mu ( \$ ) = \$ $, and, according to the equation,
$ \mu ( s ) ( \$ ) = m (s ) $). \\
\mbox{ } \hfill $\Box$\\

Note that, as in Corollary 2,
if a certain string does not belong to the language, we can assume that
the corresponding value in this table is undefined; thus $\mu$ is not
necessarily defined for all possible concatenations of strings of
$S$.\\

In view of the above theorems, any semantics is equivalent
to a compositional
semantics, and hence it would be meaningless to keep the definition of
compositionality as the existence of a homomorphism from syntax to semantics
without imposing some conditions on this homomorphism.
Notice that requiring
the computability of the meaning function will not do.
In mathematics, where semantics obviously is
compositional, we can talk about noncomputable functions, and it
is usually clear what we postulate about them. Also, we have the
following proposition. \\

{\sc {\bf PROPOSITION 4}.}
If the set of expressions $S$ and the original meaning
function $m(x)$ are computable, then so is the meaning
function $\mu(x)$.\\

Proof. One can easily check that
the table defining the meaning functions $\mu(t(\alpha))$
in the note above is effectively
computable from the functions $m(x)$ and $ t( \alpha ) $. Hence
so is the function $\mu$.
\mbox{ } \hfill $\Box$ \\

{\sc note}. What does it mean that the table is effectively
computable from the functions $m(x)$ and $t( \alpha ) $?
It means that given a Turing machine, $T1$, that prints all
elements of $S$, and another Turing machine, $T2$, that takes
an element $s$ on the output tape of $T1$ as input and produces
as output $m(s)$, we can construct a third Turing machine, $T3$, that
produces the successive elements of the table, i.e. enumerates all the
equations (e.g. for any pair $<m, n>$ gives the $n$th value pair of the
$m$th equation).
But these equations define our function $\mu$. Hence
$\mu$ is effectively computable.\\
Also, notice that the proposition holds true for generalized
computability, in the sense of \cite{Barwise75}.

\section{Systematic semantics. I. Examples}
\mbox{} \cite{Hirst87}, pp.27, talks
about compositionality, postulating that the meaning of a whole should
be a "systematic" function of the
meanings of the parts. He does not define the word
"systematic" except as being an antonym to "idiosyncratic".
What it could mean is that we want to avoid such meaning functions
as the ones defined in
the proof of Theorem 1 and the subsequent propositions.
We suggest a simple way of doing it ---
by requiring that the meaning function belong to a certain class.
By Proposition 4 this does not work if we merely postulate that
the function be computable. However it does work for
smaller classes of functions as the following examples show.
\vspace*{.1in}\\
{\sc {\bf A simple grammar without a systematic semantics}} \\
If meanings have to be expressed using certain natural,
but restricted, sets
of operations, it may turn out that even simple grammars do not have a
compositional semantics.
Consider two grammars of numerals in base 10: \\

{\bf Grammar ND} \\
    N $\leftarrow$ N D \\
    N $\leftarrow$ D  \\
    D $\leftarrow$ 0 $\mid$ 1 $\mid$ 2 $\mid$ 3 $\mid$ 4 $\mid$ 5 $\mid$
6 $\mid$ 7 $\mid$ 8 $\mid$ 9 \\

{\bf Grammar DN}   \\
    N $\leftarrow$ D N \\
    N $\leftarrow$ D \\
    D $\leftarrow$ 0 $\mid$ 1 $\mid$ 2 $\mid$ 3 $\mid$ 4 $\mid$ 5 $\mid$
6 $\mid$ 7 $\mid$ 8 $\mid$ 9 \\

{\sc {\bf PROPOSITION 5}.}
For the grammar ND, the meaning of any numeral can be expressed in the model
$( Nat, + , \ast , 10 )$ as
$$ \mu( N D ) = 10 \ast \mu(N) + \mu(D) $$
that is, a polynomial
in two variables with coefficients in natural numbers.\\
\mbox{ } \hfill $\Box$\\

On the other hand, for the grammar DN,
we can prove that no such a polynomial exists:\\

{\sc {\bf THEOREM 6}.}
There is no polynomial $p$ in two variables $x, y$ such that
$$ \mu( D  N ) = p ( \mu(D ), \mu(N) ) $$
and such that the value of $ \mu ( D  N
) $ is the number expressed by the string  $  D  N  $ in base 10.\\

Proof.
We are looking for
\begin{eqnarray*}
\mu( D N ) & = & p (\mu(D), \mu(N)) \\
& = & \mu(D) \ast  10^{length(N)}  + \mu(N) \\
\end{eqnarray*}
where the function $p$
must be a polynomial in these two variables.
If such a polynomial exists, it would have to be equal to
$\mu(N)$ for $\mu(N)$ in the interval $0..9$, and to
$ \mu(D) \ast 10 + \mu(N) $
for $ \mu(N)$ in $10..99$, and to $ \mu(D) \ast 100 + \mu(N) $ for
$\mu(N) $ in $100..999$, and
so on. Let the degree of this polynomial be less
than $ n $ , for
some $n $. Let us consider the interval
$ 10^{n} .. 10^{(n+1)}- 1$. On this interval the polynomial
would have to be equal identically
to $ \mu(D) \ast 10 ^{ n}+ \mu(N) $.
Now, if two polynomials of degrees less than $n$ agree on
$n$ different values, they must be identical. Hence,
$p (\mu(N), \mu(D)) = \mu(D) \ast 10 ^{ n}+ \mu(N) $. But this
would give wrong values for other intervals, e.g 10..99.
Contradiction. \\
\mbox{ } \hfill $\Box$\\

But notice that there is a
compositional semantics  for the grammar DN that
does not agree with intuitions:
$ \mu( D  N ) = 10 \ast \mu(N) + \mu(D) $, which
corresponds to reading the number backwards. And there are
many other semantics
corresponding to all possible polynomials in
$\mu(D)$ and $\mu(N)$.
Also observe that (a) if we specify enough values of the meaning function we
can exclude any particular polynomial; (b) if we do not restrict the degree
of the polynomial, we can write one that would give any values we want on a
finite number of words in the grammar.
The moral is that not only it is natural to restrict meaning functions to,
say, polynomials, but to further restrict them.
E.g. if we restrict the meaning functions to polynomials of degree 1,
then by specifying only three values  of the meaning function we can (a)
have a unique compositional semantics for the first grammar; and
(b) show that
there is no compositional semantics for the second grammar
(directly from the
proof of the above theorem).
\section{Some linguistic examples}

In this section we want to explain how the theorems we proved in
Section 2
apply to typical linguistic examples. In the process
we also explain how languages with operators can be handled by our
method. We begin by discussing the simple case of idioms.\\

{\sc \bf{Idioms}}.
Intuitively, the meaning of "high seas" is not compositional,
because "high" refers to length or distance, and not to open
spaces; moreover one could even argue that although "seas" is
plural, "high seas" is semantically singular, for it means
an "open sea". (However, the precise semantics of the expression
is not important at this
point). We want to show how we can assign compositional semantics
to such non-compositional examples.

Let the language $S$ consist of $ wall, seas,  high$,
$high.wall$, and $high.seas$. The equations we need to ensure
the compositionality of semantics have the familiar form:
\begin{eqnarray*}
\mu(seas) & = & \{ < seas, m(seas)> \} \\
\mu(wall) & = & \{ < wall, m(wall)> \} \\
\mu(high) & = & \{ < high, m(high)> ,
< \mu(wall), \mu(high.wall)> , \\
          &   &  < \mu(seas), \mu(high.seas)> \} \\
\mu(high.seas) & = & \{ < (high.seas) , open(m(sea))>\} \\
\mu(high.wall) & = & \{ < (high.wall) , high(m(wall))>\} \\
\end{eqnarray*}
Notice, that we could add $building$ and other words
to the language and easily
extend this set of equations. The intuition we associate with
compositionality would be captured by the uniformity of
the meanings of $high.X$ as $high(m(X))$, where $X$ ranges over
$wall, building, ...$.
However the formal expression
of this intuition as the principle of compositionality does not work,
which can be seen by noticing that the meaning of $high.seas$ is
a composition of the meaning functions for $high$ and $seas$.
What is happening has to do with the fact that, by
definition, functions defined by cases are as good as any others.
And what we have done is to have defined the meaning of $high$ by
cases.\\

{\sc \bf{Coordination}}.
We now turn our attention to a slightly more complicated example.
Consider disjunction and conjunction, $+$
and $\&$.  We plan to prove the following results:\\

{\sc {\bf PROPOSITION 7}.} Let $+$ and $\&$ denote "or" and "and".
Then: \\
  (A). It is possible to assign compositionally the "natural"
semantics of $(a+b)\&c$
to expressions of type $a+(b\&c)$ and preserve the original
meanings of $a+b$ and $b\&c$. \\
  (B). (A) is not possible if the meaning functions have to be Boolean
polynomials. \\
\vspace*{.1in}\\
Proof.
To keep things as simple as possible, consider language $S$ consisting of
$a+(b\&c)$, $b\&c$, $b$, $c$, and $a$. To apply directly the
solution lemma we should represent the operators in their prefix form;
e.g. $a+b$ becomes $+.a.b$. Then our language $S$ becomes
$+.a.\&.b.c$, $\&.b.c$, $+$, $\&$, $b$, $c$, and $a$. (And for the sake
of completeness we can add to it $a.\&.b.c$, and $b.c$).
As before we write our equations (this time using the version
with \$):
\begin{eqnarray*}
\mu(+.a.\&.b.c) & = & \{ < \$ , (m(a)+m(b))\&m(c)> \} \\
 \mu(a.\&.b.c) &=& \{ < \$ , <a,m(b)\&m(c)>> \} \\
\mu(\&.b.c) & = & \{ < \$ , m(b)\&m(c)>   \} \\
 \mu(b.c) &=& \{ < \$ , <m(b),m(c)>>   \} \\
 \mu(b) &=& \{ < \$ , m(b)>, < \mu(c), \mu(b.c)>   \} \\
 \mu(\&) &=& \{ < \$ , m(\&) > , <\mu(b.c), \mu(\&.b.c)>  \} \\
 \mu(a) &=& \{ < \$ , m(a) > , <\mu(\&.b.c), \mu(a.\&.b.c)> >  \} \\
 \mu(+) &=&
\{ < \$ , m(+) > , <\mu(a.\&.b.c), \mu(+.a.\&.b.c) > , \\
 & &  <\mu(a.b), m(+.a.b) > \} \\
\end{eqnarray*}
It can be easily checked that as before $\mu(s.t)=\mu(s)(\mu(t))$ for all
$s , t \in S$. The meaning $m(a)$ of $a$ is arbitrary, but we
would typically identify it with its logical value (true or false).
Also, notice that without loss of generality those  $a, b, c$
(and hence the language $S$)
can be "expanded" to sets of variables
 $a_i , b_j , c_k $, resulting in a slight change in the equations,
but not changing the content of the theorem. Then,
for all pairs of variables disjunction and conjunction would behave as
usual; however for a combination of a variable with a
Boolean formula they could behave arbitrarily.
This proves the first part.\\

We now prove the second part, i.e. that if the meaning functions are
restricted to Boolean polynomials, it is impossible to assign
compositional semantics given by the "natural" semantics of
$(a+b)\&c$
to expressions of the type $a+(b\&c)$ and preserve the original
meanings of the connectives. (The "natural" semantics is of
course the conjunction of the Boolean value of $a + b $ with the
Boolean value of $c$).

The proof consists in observing that
$ (m(a) + m(b)) \&  m(c) $ cannot be obtained as a Boolean polynomial
(function) $ p( m(a) , m(b)\&m(c)) $. To see it, it is enough to
construct a truth table showing the values of $(a+b) \& c$ and
$ b \& c $  and observing that there cannot be a functional
dependence of the former on the latter and the value of $a$
(compare the values for the triples $< a=1 , b =0 , c=1 >$ and
$ < a=1 , b =1 , c=0 >$).
\mbox{ } \hfill $\Box$\\

\section{Systematic semantics. II. Discussion}
Above, we have argued that the existence of a homomorphism from
syntax to semantics does not restrict the grammar, but if we put
some constraints on such a homomorphism, they actually might restrict
grammars of languages. We have called such
homomorphisms {\em (F-)systematic}. However the nature of systematicity
seems to be very much an open problem. In this section we discuss
some of the most obvious issues, and propose some research
possibilities in this area.

The first
natural question that arises is: What should be this class $F$? We have
shown that for a grammar of numbers, and a grammar of two Boolean
connectives the natural classes are polynomials. Clearly, this cannot
always be the case. For instance, it seems natural to map verbs
into predicates and nouns into their arguments. But we know that
if we want to provide compositional semantics for more than the
simplest case of subject-verb-object construction we need other
mechanism, e.g. type raising. On the other hand, unrestricted type
raising leads to the results we have just discussed. We arrive then
at the following variant of the natural question: How should we restrict
type raising? (So that we can account e.g. for ellipsis, but at the
same time constrain the grammars).

Many grammatical constructions express meanings that go beyond
expressing predicate-arguments assignments. For example, "the X-er,
the Y-er" construction
(\cite{Fillmal88}, \cite{cons1}), as in "the more you
dive, the better you swim", expresses a proportional dependence.
Other constructions can express speech acts, various kind of conflict,
etc., hence creating rich sets of meanings. The next question we
can ask is whether we can express constraints on the syntax-semantics
interface by saying that the homomorphism should be simple,
relative to the the class of
meaning functions. A mathematical analogy would be
to say that we are given
many different functions, e.g. $sin(x), cos(x), e^x, ...$,  but
only simple ways of composing them, e.g. only by the $+$.

In essence, when we say that a function is {\em  F-systematic}, we view
{\em F} as a measure of complexity and/or expressive power.
Hence the natural association with polynomials. But there are other
measures of complexity.
\cite{Savitch93} discusses grammars and languages in terms of
Kolmogoroff complexity, suggesting that more compact grammars
are better even if they overgeneralize (e.g. by approximating
a finite language by an infinite one).
We believe that his work is relevant for systematicity, but
we do not have any formal argument supporting that claim, except
the observation that polynomials are more compact than functions
defined by cases. So perhaps this might be a beginning of a formal
connection.

One of the referees has pointed out two other ideas. First,
there could be other more natural notions of systematicity, in cases
when meanings are specified
by means of constraint solving, as is implicit in unification-based
formalisms, and even in Theorem 1, where meanings are extracted as
solutions to equations. The second idea,
the differences between natural and formal languages notwithstanding,
is a programming language semantics
concept which actually restricts the ability of a semantics to be
compositional. This concept is "full abstraction", i.e. the
equivalence between the operational and denotational semantics,
(cf. \cite{Gunter93}), which can be viewed as a general constraint on
compositionality.

A radically different approach to the interaction of syntax and semantics
has been presented in \cite{cons1} and \cite{Jurafsky92}.
Language is modeled there as a set of constructions (cf. also
\cite{Fillmal88} and \cite{Bloomfield33}).
In that model there is no separate syntax, since constructions encode both
form and meaning. Each construction
explicitly defines the meaning function taking the meanings of
its subconstructions as arguments.
Intuitively, in that model we assume that
each word sense requires a separate semantic description;
the same is true about each
idiom, open idiom (\cite{Fillmal88}), and a phrasal construction.
This means that we make each
semantic function as complicated as linguistically
necessary, but their mode of combination is restricted.
(Continuing the above mathematical analogy, we would say that
the only mode of combination is substitution for an argument).
In the construction-based model semantics is "compositional"
and "systematic" (with respect to the set of all these
meaning functions), but there is no homomorphism from syntax
to semantics, because there is no syntax to speak of.
It is "compositional", because the meaning of a construction is
a function of the meanings of its parts and their mode of
combination. (Note that such a function
is different for different constructions, and each construction
defines its own mode of combination). And it is systematic, because
the modes of combination are not arbitrary, as they have to be
linguistically justified.
But since only few formal aspects
of constructions have been worked out, we can only speak of that model
as of yet another possibility.

The last point we want to make is that while it is true that the
homomorphism condition is too weak (in general) to count as
systematicity, the semanticists (e.g. Montague)
have not been using arbitrary homomorphisms. Thus a careful
examination of their work should lead to some characterization
of "good" homomorphisms; and this seems to be another interesting
avenue of research. (As suggested by one of the referees, a technique
from universal algebra might also prove helpful,
where one first gives a class
of algebras and then specifies meanings as homomorphisms
from the initial algebra of the class).

\section{Conclusions}

In this paper we have shown the formal vacuity of the compositionality
principle. That is, we have shown that the property that the meaning
of the whole is a function of the meanings of its parts does not put any
material constraints on syntax or semantics.
Theorem 1 (and its corollaries) explain formally why
the postulate of a
homomorphism between syntax and semantics is not restrictive
enough: the syntactic operator of concatenation
"$.$" can be mapped into functional
composition operating on functions that encode arbitrary semantics
of an arbitrary language.
As we have seen in the examples, the
presence of other operators does not change the result, because
they can be treated as yet another letter of the alphabet, and
one can still produce a homomorphism between syntax and semantics.

The problem of the vacuity of compositional semantics
arises, because in the formal definition of compositionality
meaning functions can be completely arbitrary. Therefore we have
proposed that the meaning functions should be systematic, i.e.,
non-arbitrary.
We have shown that this notion makes sense formally; that is,
we presented examples
of semantic classes of functions, for which there are grammars with
meaning functions in that class, as well as we have shown that there
are grammars that cannot have a meaning function in that class.
As we noted in the previous section, both
the formal and the linguistic nature of systematicity
remains an open problem, but with a few promising avenues of research.

In this paper we have restricted ourselves to the mathematics of
compositionality, and many important issues have not been discussed.
For instance, the main result is relevant for
theories of grammar and for the thesis about
the reduction of syntax to lexical meanings (cf. e.g.
T. Wasow on pp.204-205 in \cite{Sells85}). Also,
systematicity of semantics should give us a handle in
constraining the power of the semantic as well as the syntactic
components of a grammar (cf. \cite{systsem}).
Furthermore, our results have implications for
computational linguistics (they are briefly discussed in
\cite{Zad92coling}).

Finally, the reader should note that one of the more bizarre
consequences of Theorem 1 is that we do not have to start building
compositional semantics for natural language beginning
with assigning of the meanings to words. We can do equally well by
assigning meanings to {\em phonems}  or even {\em LETTERS}, assuring
that, for any sentence, the intuitive meaning we associate with
it would be a function of the meanings of
the letters from which that sentence is composed. But then
the cabalists had always known it.

{\sc {\bf Acknowledgements}}. I would like to thank Alexis Manaster
Ramer for our many
discussions of compositionality, and the referees for their
insightful comments.

\end{document}